# Modelling of Walking Humanoid Robot With Capability of Floor Detection and Dynamic Balancing Using Colored Petri Net

Saeid Pashazadeh and Saeed Saeedvand

Faculty of Electrical and Computer Engineering, University of Tabriz, Tabriz, Iran

**ABSTRACT**

*Most humanoid robots have highly complicated structure and design of robots that are very similar to human is extremely difficult. In this paper, modelling of a general and comprehensive algorithm for control of humanoid robots is presented using Colored Petri Nets. For keeping dynamic balance of the robot, combination of Gyroscope and Accelerometer sensors are used in algorithm. Image processing is used to identify two fundamental issues: first, detection of target or an object which robot must follow; second, detecting surface of the ground so that walking robot could maintain its balance just like a human and shows its best performance. Presented model gives high-level view of humanoid robot's operations.*

## 1. INTRODUCTION

Humanoid robots are group of robots that robot designers have long desire to make them more similar to humans such that they could replace them instead of people for doing different works in harsh environments like transportation of nuclear devices. One of the most important issues in making a robot is keeping its balance when it stands or moves. When a robot is standing or walking, it can maintain its balance on uneven surfaces using different sensors like Gyroscope and Accelerometers that are controlled by various algorithms that examples of which are given in [1- 4]. Main problem is that usually robot has little information about floor on which it is moving, therefore it can often detect the surface and show appropriate response only after it has reached the intended place. This usually gets the robot into trouble, disturbs the balance of the two-leg robot and sharply decreases its speed. This paper aims to present an algorithm by which the robot detects the floor, follows a predetermined target, and shows the desired reactions.

Section 2 of paper gives a brief explanation about modelling using Colored Petri Nets (CPN) [5-7] and then addresses the operation of balance sensors as well as the proposed algorithms for noise filtering. Later, the proposed algorithm is modelled using CPN-Tools software. In the image processing part of presented model using CPN, different filters applied to captures images of cameras and the way in which image processing is used to detect the floor are investigated. Finally, the proposed method is implemented and tested, and the results of applying it to the designed robot (SoRoBo) are analyzed.

Many projects have been carried out on the humanoid robots, each having its own ideas and methods. Some of these projects are as follows: Jung-Yup Kim et al. developed a control algorithm in which they controlled their robot on a sloping ground with little roughness, using control sensors in dynamic and stable states [1]. Sebastien Dalibard et al. presented an interesting two-stage method to control the robot's dynamic walking through narrow spaces on a smooth ground [2]. In their method, Andrew L. Kun and W. Thomas Miller making use of an Accelerator sensor upon nervous systems for keeping balance of walking robot [3]. Kenji Kaneko et al. introduced the platform of the Japanese robot (HRP-2) which is able to move on rough surfaces at two speeds and on an even line using Gyro and Accelerator sensors [4]. Karungaru et al. developed an algorithm for determining position of their robots using image processing. By real-time image processing and filtering of image blur during movement of robot, they focused on using image processing to control walking humanoid robot. They were sending results of image processing to robot for immediately controlling it [8]. Jung-Yup Kim et al. conducted an experiment on their robot, in which they first set a target in stereovision and controlled their robot's movement on an even ground towards the target [9]. Hirai, K et al., too, presented a control algorithm for Honda robot that can do various movements and sustain its balance [10].

## 2. COLORED PETRI NET

Petri Net [5-7] is a modelling method that benefits from graphical representation for analysis of activities involved in synchronous and concurrent systems. A Petri net consists of four basic elements: places, transitions, arcs, and tokens. Assignment of tokens to places (markings) represents states of the system. A place is shown in the form of a circle or oval and tokens in the form of small filled inside circles that resides in the places. Each transition is shown in the form of a rectangle and represents system activities. Petri nets works based on its special enabling and firing rules.

Many extensions to classical Petri net is developed that aims of all of them are extending modelling capability of Petri nets. Colored Petri net is the most recent and powerful extension of classical Petri net that enables modeller to define colour type (data type) for tokens and their containing places. Its modelling capability is extended using ML programming language that is an artificial intelligence language. ML allows using inscriptions and functions as arc expressions and guard conditions of transitions. Using colored Petri net enables modeller for modelling wide range of systems. ML language that is used in colored Petri net is revision of original one that some futures of it is removed and some new futures are added to it that make it synchronized with Petri net terms such as defining multi-set operators and multi-set markings [11],[12]. One of the best tools that is developed for modelling and analysis of colored Petri net models is the CPN Tools. This free open source software is accessible from its web site [13]. Colored Petri net can be used for modelling and verification of wide range of protocols and rules in different fields of computer science like security and database systems [14],[15]. Modelling of humanoid robot using colored Petri net is under study in this paper.

## 3. GYROSCOPE AND ACCELERATOR SENSORS

Gyroscope and accelerator are most commonly used sensors for controlling majority of dynamic walking humanoid robots. Gyroscope sensors are used to measure direction of robot by measuring yaw, pitch and roll angles of it and accelerator sensors are used for measuring motion gradient in (x, y, z) axes directions in real-time manner. Proposed algorithm in this paper uses both of these sensors. When humanoid robot is moving, its skeleton suffers from violent convulsions that cause noise and inaccuracy of data read by sensors. Kalman Filtering algorithm is often used to overcome this problem and yields desired results. Method that was presented by Ferdinando et al. [16] is used to obtain some filtered data from two different coordinates which are likely to bring about noise. Performance of these two sensors is shown and is explained by model provided in section Gyroscope and Accelerometer.

## 4. MODEL OF HUMANOID ROBOT

A hierarchical model of a humanoid robot is present in this paper. Figure 1 shows top-level model of system. Model contains three substitution transitions. Description of each substitution transition is as follows:

**Gyroscope and Accelerometer Transition:** Sub module of this substitution transition is responsible for producing filtered coordinates after noise elimination in x, y, and z axes from control sensors, indicating correct acceleration in specified small time spans. Output of this transition transmits to place *Filtered Coordinate* with colorset of type *f_coordiante* that is some kind of list.

   *colset axis= int with 1..330;*
   *colset filtered_coordinate=product axis*axis*axis;*
   *colset f_coordinate=list filtered_coordinate;*

List elements are of colorset product of three variables of colorset axis that shows correct gradient of each angle. Sub model of substitution transition *Gyroscope and Accelerometer* is appeared as *Tilt sensor*, which will be explained later in detail in section 4.1.

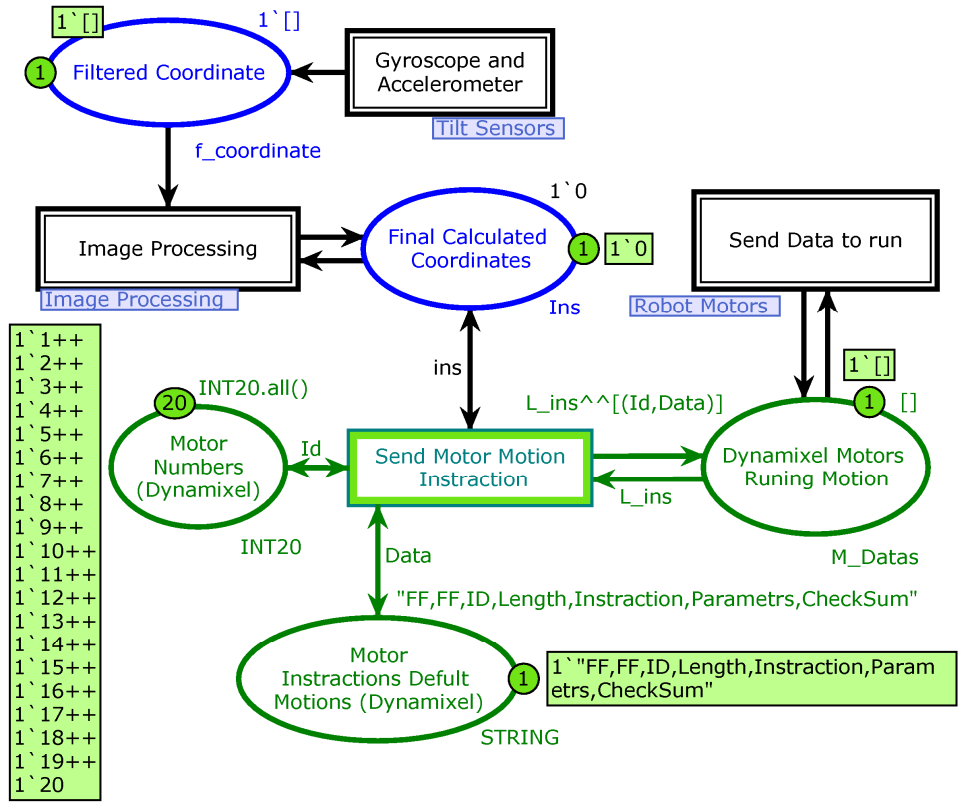

Figure 1. Top-level model of humanoid robot's operation

**Image processing Transition:** Sub module of this substitution transition carries out operations that are related to image processing. Transition fires processed information at place *Final Calculated Coordinate*. Sub module of this substitution transition is named as Image Processing and will be addressed in section 4.2. Transition *Send Motor Motion Instruction Transition* is responsible for receiving data from following places:

> **Final Calculated Coordinates place:** Colorset of this place is of type int. This place specifies instruction code for determining control operations of robot. Amount of this code is obtained in subpage that will be explained in section 4.2, and determines which control instruction has been calculated in *Image Processing* subpage.
> **Motor Numbers (Dynamixel) Place:** Colorset of this place is of type string and specifies control package for controlling each of engines, regarding motions that already have been made based on mechanical structure of the robot. Here, protocol Dynamixel is used for Dynamixel-RX28 engines [17].

**Send Motor Motion Instruction Transition:** This Transition, after binding required data from places that are connected to it, sets out to calculate instructions which must be binding to each of engines based on the control instruction involved in place *Final Controlling Instruction and Motion* from place *Motor Instructions Default Motions (Dynamixel)* by variable Data. Then, it changes instruction, engine ID and other required quantities existing in protocol of Dynamixel engines and fires them into place *Dynamixel Motors Running Motion* to send them to the engines.

### 4.1. Tilt Sensors Sub Module

Figure 2 shows subpage of Tilt Sensors sub module of substitution transition *Gyroscope and Accelerometer* substitution of Figure 1. This sub model contains 2 transitions and 6 places that responsibilities of them are as follows. Place *Gyroscope Sensor* holds change in rotation of humanoid robot around x, y, and z axis respectively and place *Accelerometer Sensor* holds acceleration of robot in direction of three axes relative to humanoid robot in each step. We assumed in model that sensed information of gyroscope and acceleration sensors are placed in these places. We generated random values as sensed data for these sensors in the model. Accelerometer indicates correct gradient of robot at each moment (x, y, z), and Gyroscope measures direction or turn of robot at each moment (yaw, pitch, roll). Some of color sets that are used for modelling these two sensors are as follows:

*colset Angle = int with 1..330;*
*colset Coordinate_g = record yaws : Angle * pitchs : Angle * rolls : Angle;*
*colset Acc = int with 1..330;*
*colset Coordinate_ac = record x_ac : Acc * y_ac : Acc * z_ac : Acc;*

Color set *Angle* is defined of type sub range 1..330 of int for representing directional change of robot along each axes. Color set *Coordinate_g* is a record of three fields that each one represents amount of robot's rotation along each of three axes. Color set *Acc* is defined of type sub range 1..330 of int for representing acceleration of robot along each axis. Color set *Coordinate_ac* is a record of three fields that each one represents amount of robot's acceleration along each of three axes.

Transition *Gyroscope and Accelerometer* continually binds produced coordinates by each sensor from places *Accelerometer Sensor* and *Gyroscope sensor* and puts them in places *Accelerometer Data* and *Gyroscope Data*. places *Accelerometer Sensor* and *Gyroscope Sensor* must not considered as real sensors, they present result of sensing and pre-processing of their sensed data. *Place Number* causes that after reading of sensors inputs transitions *Gyroscope and*

*Accelerometer* becomes disabled until filtering of these information will be done. Transition *Kalman Filtering* uses data from places *Accelerometer Data* and *Gyroscope Data* using Kalman Filtering algorithm proposed by Ferdinando [16] and sends result in place *Filtered Coordination*. Operation of transition *Kalman Filtering* is demonstrated by making average of coordinates of two sensors in the Transition. More information about Kalman filleting is available in [18].

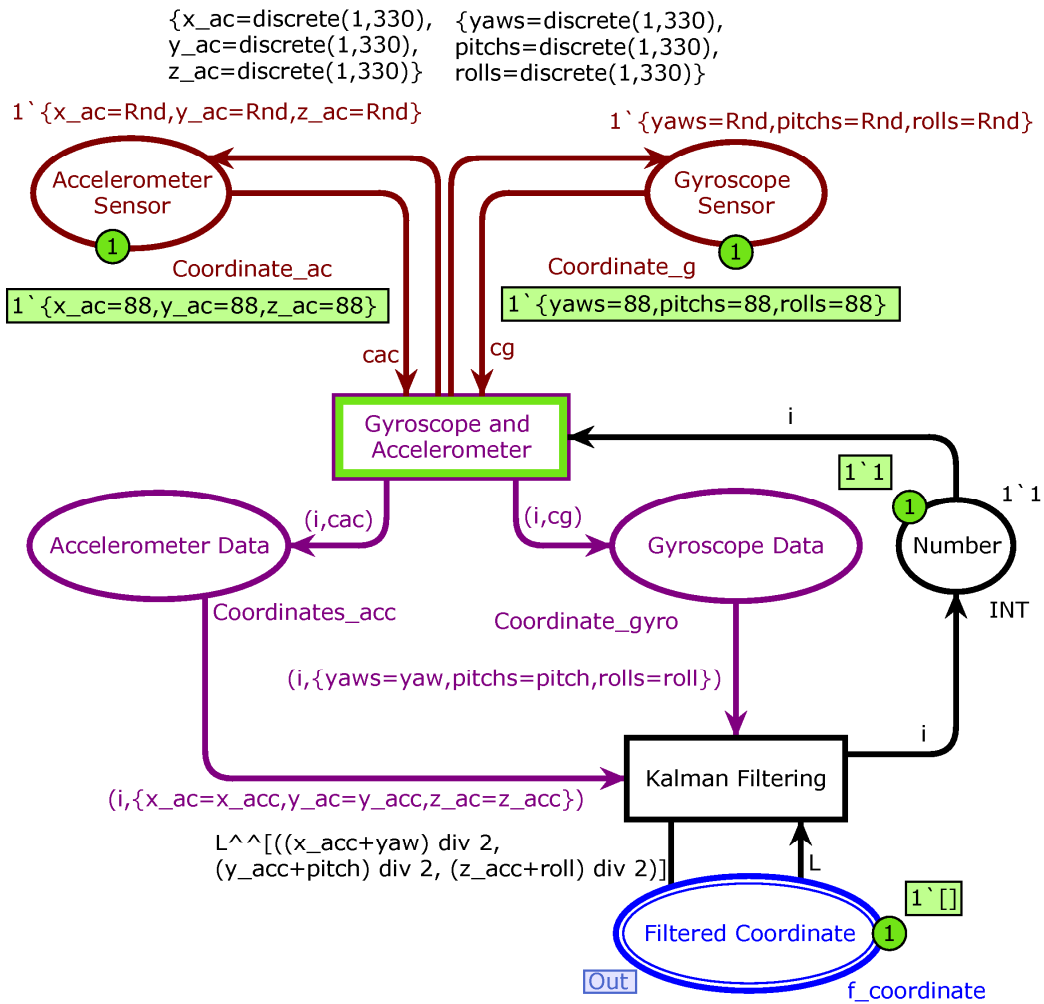

Figure 2. Gyroscope and Accelerometer subpage

### 4.2. Image Processing Sub Module

Figure 3 shows *Image Processing* sub module of substitution transition *Image Processing* in Figure 1. As name of transition implies, operations of image processing are carried out in this part. Place *camera* is regarded as image processing camera, in which every frame of the taken pictures is stored. When the humanoid robot is walking, small movement of robot shakes robot's camera and make blurred take images. To overcome this problem, operations of image stabilization in transition *Anti-Image Vibration Filtering* are carried out. Task of this transition is to clear up blurred vision when robot is walking. Let Assume that method proposed by J. Windau et al. [19] was used in implementation of operations. Real-time video image stabilization system (VISS) was implemented in several layers that primarily is developed for aerial robots. Its unique architecture combines four independent stabilization layers.

International Journal in Foundations of Computer Science & Technology (IJFCST), Vol.4, No.2, March 2014

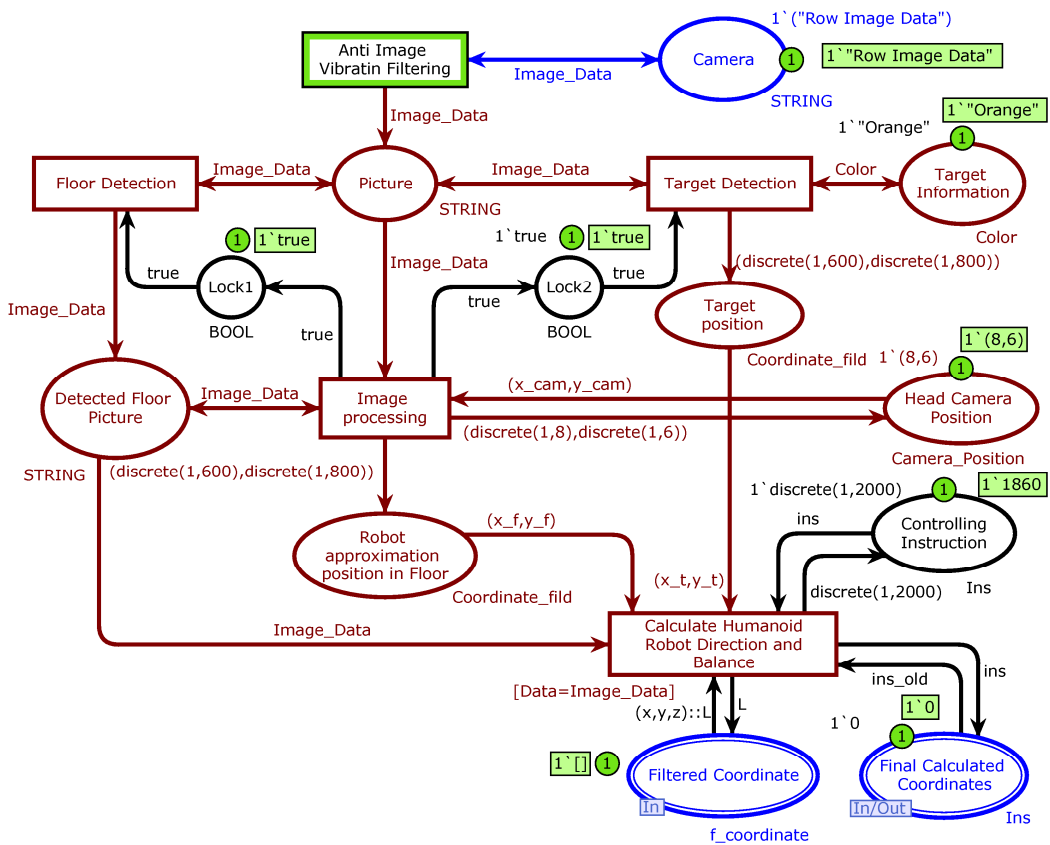

Figure 3. Image processing subpage

First layer detects vibrations via an inertial measurement unit (IMU) and performs external counter-movements with a motorized gimbal. Second layer damps vibrations by using mechanical devices. Internal optical image stabilizes image of camera in third layer and finally, fourth layer filters remaining vibrations using software. This study is not aimed at detailing this implementation due to enormous volume of information. In general, VISS operations are considered as *Anti Image Vibration Filtering* transition in our model. Then anti-image vibration filtering is applied to the image and new image will be located in place *Picture*. In this stage, each of two following transitions on image must be run once and this is controlled by places *Lock*. First transition is *Target Detection*, whose task is finding intended target on image with regard to the information that exists in place *Target Information*. Take an orange ball as an example, if intended target is spotted from the image in that frame, coordinates of target on image are sent to place *Target Position* and in otherwise, zero quantity is sent to this place. Second transition is *Floor Detection*, which is a very important and has responsibility of detecting floor. Young-geun Kim et al., in a study conducted in 2004 introduced an interesting method for detecting floor by means of various filters [20]. Let assumed that their method is used in our model of humanoid robot system. Transition *Floor Detection* binds new image from place *Picture* and after conducting necessary operations, detects image of floor and sends them to place *Detected Floor Picture*.

Transition *Image Processing* will be enabled after firing of these two transitions. Let assume that place *Detected Floor Picture* contains taken image of the floor and place *Head Camera Position* contains current position of robot's camera that amount of it is in form of two dimensional coordinates in range of (x=(1..8), y=(1..6)). After binding of input arc inscriptions of transition *Image Processing* fires and calculates approximate position of the robot and sends them at place *Robot Approximate Position in floor*.

Last transition of this subpage is *Calculate Humanoid Robot Balance and Way*. It is responsible for processing balance instructions and calculating path of the robot through binding existing data in places *Target Position*, Robot *Approximate Position in floor*, *Controlling Instruction*, and *Filtered Coordinate*. Tokens of place *Controlling Instructions* are instructions of control unit for movement of robot. This transition uses tokens of mentioned places and by firing computes next position of robot and sends them to place *Final Calculated Coordinates*. Values of control instructions are considered random values from uniform distribution in this model.

### 4.3. Robot Motors Subpage

Figure 4 shows *Robot Motors* subpage that includes detailed model of substitution transition *Send Data to run* in Figure 1. This sub model contains 20 places that each one is responsible for controlling one part of engine of humanoid robot that is shown in Figure 5. Transition *Send instruction to Motors (Dynamixel)* takes incoming instructions that are stored in place *Dynamixel Motor Running Instruction* then translates and sends them to appropriate motor of robot's engine such that each engine can carry out its instructions, causing the robot to respond.

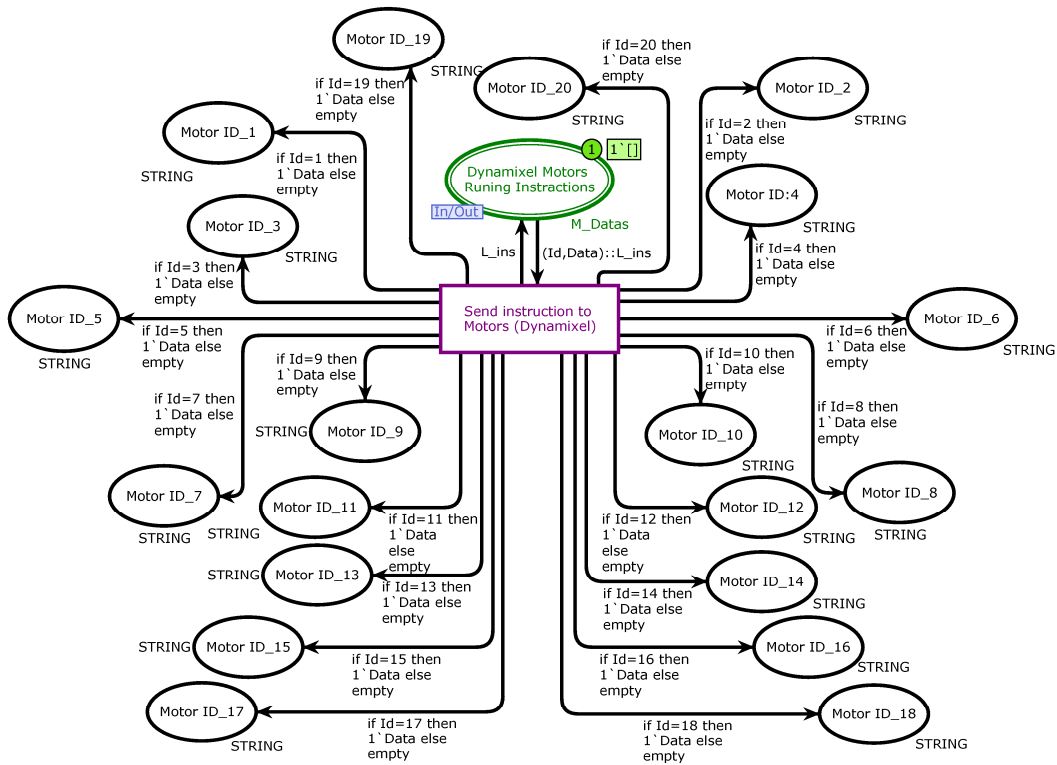

Figure 4. Robot motors subpage

## 5. RUNNING PROPOSED MODEL OF SOROBO HUMANOID ROBOT

Figure 5 shows SOROBO humanoid robot that is 54 centimetres tall. Figure shows joints and places of robots motors and their degree of freedom. Structure of this robot is used in modelling proposed algorithms in this paper.

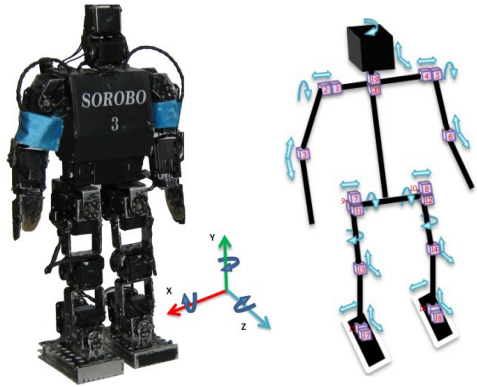

Figure 5. SOROBO robot and its motors structure

Experiment done on the robot has yielded good results so that the robot is able to detect assigned target without getting out off the floor and continue walking dynamically without conflict with any obstacles. Robot has a camera with two moving angles with engines No. 19 and No. 20, marked with their movement angles in Figure 5. In this robot, Dynamixel engines of RX28 series have been used. Inside main body of robot, there is a RoBoard RB-110 for controlling its different parts. Robot has 20 engines, and as mentioned earlier, engines 19 and 20 control movements of robot's head in two different directions. Main engines whose amounts are changed to maintain robot's balance in Real-time manner are engines 1 to 6 at upper body and engines 7 to 18 at lower body. These engines, during robot's walking controlled by motions designed for different movement directions, must be changed by transition 'Calculate Humanoid Robot Balance and Way' so robot's balance is kept in different conditions.

## 6. CONCLUSION

Given enormous complexity of humanoid robots and limitations in explaining all of its parts in detail, a general algorithm was proposed for humanoid robots that enables robot to automatically detect floor, dynamically find a preset target on a sloping and uneven surface and walking towards it. Proposed algorithm was modelled using colored Petri net and presented by CPN-Tools for more future study on the model. This model presents workflow of humanoid robot in high level abstraction. Model checking shows correct operation of robot and can be used for verification of different aspect of robot's operations. This model planned to be used for performance evaluation of robot and future studies.

**Authors**

**Saeid Pashazadeh** is Assistant Professor of Software Engineering in Information Technology Department at Faculty of Electrical and Computer Engineering in University of Tabriz in Iran. He received his B.Sc. in Computer Engineering from Sharif Technical University of Iran in 1995. He obtained M.Sc. and Ph.D. in Computer Engineering from Iran University of Science and Technology in 1998 and 2010 respectively. He was Lecturer in Faculty of Electrical Engineering in Sahand University of Technology in Iran from 1999 until 2004. His main interests are modelling and formal verification of distributed systems, computer security, wireless sensor/actor networks, and applications of artificial neural networks. He is senior member of IACSIT and member of editorial board of journal of electrical engineering at University of Tabriz in Iran.

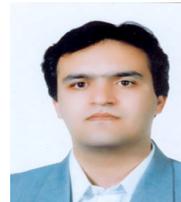

**Saeed Saeedvand** is Ms.c student of software engineering in Faculty of Electrical and Computer Engineering in University of Tabriz in Iran. He is working as a Lecturer in Islamic Azad University of Iran. He is Capitan of SoRoBo Humanoid-Kid size robotic team in Islamic Azad University and he has two champions in IRANOPEN 2011 and 2012 games. His research interests are Robotic and artificial intelligence.

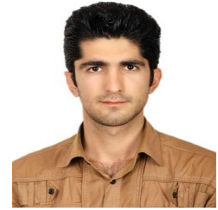